%% file: arxiv.tex
\documentclass[runningheads]{llncs}

\usepackage{eccv}

\usepackage{eccvabbrv}

\usepackage{graphicx}
\usepackage{booktabs}

\usepackage{tabularx}
\usepackage{xr-hyper}
\input{preamble}

\definecolor{cvprblue}{rgb}{0.21,0.49,0.74}
\usepackage[pagebackref,breaklinks,colorlinks,allcolors=cvprblue]{hyperref}

\usepackage{xcolor}
\definecolor{figure_red}{HTML}{e03131}
\definecolor{figure_green}{HTML}{2f9e44}

\usepackage[accsupp]{axessibility}  

\usepackage{hyperref}

\usepackage{orcidlink}

\makeatletter
\newcommand{\printfnsymbol}[1]{%
  \textsuperscript{\@fnsymbol{#1}}%
}
\makeatother

\begin{document}

\title{PuzzleCraft: Exploration-Aware Curriculum Learning for Puzzle-Based RLVR in VLMs} 

\titlerunning{PuzzleCraft: Exploration-Aware Curriculum for RLVR in VLMs}




\author{Ahmadreza Jeddi\inst{1,2,3}\thanks{Equal contribution} \and
Hakki C. Karaimer\inst{1}\printfnsymbol{1} \and
Hue Nguyen\inst{1} \and
Zhongling Wang\inst{1}\thanks{Equal contribution} \and
Ke Zhao\inst{1}\printfnsymbol{2} \and
Javad Rajabi\inst{1,2,3}\printfnsymbol{2} \and
Ran Zhang\inst{1}\printfnsymbol{2} \and
Raghav Goyal\inst{1} \and
Konstantinos G. Derpanis\inst{1,3,4} \and 
Babak Taati\inst{2,3} \and
Radek Grzeszczuk\inst{1}}

\authorrunning{A.~Jeddi et al.}

\institute{AI Center-Toronto, Samsung Electronics \and
University of Toronto \and
Vector Institute \and
York University\\
\email{ajeddi@cs.toronto.edu, hakki.k@samsung.com}}

\maketitle

\input{sec/0_abstract}

\input{sec/1_intro}
\input{sec/2_related_work}

\input{sec/3_method}
\input{sec/4_exps}

\input{sec/5_conclusion}

\bibliographystyle{splncs04}
\bibliography{main}

\appendix
\input{sec/X_supp}

\end{document}

%% file: preamble.tex
\usepackage{graphicx}
\usepackage{subcaption}
\usepackage{multirow}
\usepackage{xcolor,colortbl}
\usepackage{algorithm}
\usepackage{algpseudocode}
\usepackage{tikz}
\usepackage{graphicx}
\usetikzlibrary{spy,backgrounds}
\usepackage{soul}
\usepackage{booktabs}

\usepackage{tikz}

\usepackage{enumitem}

\usepackage{adjustbox}


%% file: sec/0_abstract.tex
\begin{abstract}
RL post-training with verifiable rewards (RLVR) has become a practical route to eliciting chain-of-thought reasoning in vision--language models (VLMs), but scaling it in the visual domain remains challenging due to costly or noisy supervision and reliance on external verifiers. Puzzle-based RLVR is a promising alternative, yet existing approaches often treat puzzle rewards as flat or sparse, which weakens group-relative learning signal. Existing curriculum strategies are overly restrictive: they rely mainly on reward statistics and do not account for exploration in the solution space, which can lead to collapsed rollout dynamics. Further, RL post-training can induce reasoning--answer inconsistency as training progresses. To address these shortcomings, we present \textbf{PuzzleCraft}, a supervision-free framework that scales vision-centric RLVR using a set of lightweight puzzle environments with built-in verification. PuzzleCraft instantiates three puzzles inspired by classic visual pretext tasks: PatchFit, Rotation, and Jigsaw. We introduce a curriculum that combines difficulty with an exploration signal derived from solution-space dispersion, and use it to downweight collapsed prompt groups. In addition, we introduce a new post-training metric, Reasoning-Answer Consistency (RAC), to measure the degree that the chain-of-though supports the answer, and show our exploration-aware curriculum improves RAC and downstream performance. Across a broad suite of vision-centric benchmarks, PuzzleCraft improves robustness and reasoning consistency, yielding consistent downstream gains on both Qwen2.5-VL and Qwen3-VL backbones. Overall, our results suggest that scalable puzzle-based RLVR benefits from curricula that account for both difficulty and solution-space collapse, together with explicit consistency-enhancing schemes. Project page: {\tt\small \url{https://puzzlecraftgrpo.github.io/}}

\keywords{VLMs \and RLVR \and Curriculum \and Reasoning consistency}
\end{abstract}

%% file: sec/1_intro.tex
\section{Introduction}
\label{sec:intro}

Recent progress in vision--language models (VLMs) has been driven by RL post-training, which moves beyond supervised instruction tuning and offers a practical way to elicit stepwise reasoning behaviors \cite{huang2025vision, chen2025sft-or-rl, zhang2025r1-vl, wang2025vl}. Among these approaches, GRPO-style objectives for RL with verifiable rewards (RLVR) have gained traction \cite{shao2024deepseekmath, guo2025deepseek, liu2025visual, shen2025vlm-r1}. Yet scaling GRPO-style RLVR in the visual domain is bottlenecked by a practical constraint: obtaining vision-centric, reliably verifiable reward signals is costly and noisy, and often requires curated supervision or external verifiers \cite{chen2025selfquestioninglanguagemodels, wu2025visual-jigsaw}. This has motivated growing interest in puzzle environments, self-contained visual tasks with automatic verification, as a promising direction for scalable, supervision-free RLVR \cite{noroozi2016unsupervised, misra2020self, wang2025jigsaw-r1, wu2025visual-jigsaw}.

Existing puzzle-based RLVR pipelines leave several gaps that limit how far this direction can go. First, many works focus on a single puzzle in isolation, making it unclear whether gains reflect general improvements in multimodal reasoning or narrow specialization; inter-puzzle transfer and generalization remain under-explored. Second, puzzle RLVR often relies on binary success rewards, which are sparse and do not reward partial progress. Third, rewards are frequently treated as flat across difficulties, so easy, medium, and hard instances contribute with nearly the same influence on updates \cite{zhang2025grpoleaddifficultyawarereinforcementlearning, zhou2025darodifficultyawarereweightingpolicy}. This can exacerbate vanishing-advantage dynamics when rollouts within a group become homogeneous, driving group-relative advantages toward zero and weakening learning \cite{wang2025vl}. Moreover, the existing curriculum solutions in VLM GRPO only consider the reward statistics without considering the rollout exploration, which as we observe, limits the gains. Relatedly, existing curriculum strategies for VLM GRPO~\cite{guo2025observe-r1, jiang2025vcrlvariancebasedcurriculumreinforcement} typically rely only on reward statistics and do not account for whether rollouts continue to explore or instead collapse into repetitive, low-diversity behavior. As we observe, this omission can cap achievable gains. Finally, puzzle training exposes a broader RLVR failure mode: as post-training progresses, models may increasingly shortcut or drift into reasoning--answer inconsistency, where the chain-of-thought (CoT) no longer supports the final answer \cite{huang2025answer-consistent, chen2025grpo-care, yao2025reasoningmodelspronehallucination, chen2025reasoningmodelsdontsay}. Together, these gaps suggest that unlocking the full potential of puzzle-based RLVR requires richer reward signals, curricula that account for both difficulty and rollout collapse, and explicit attention to consistency during optimization.

We present \textbf{PuzzleCraft}, a supervision-free RLVR framework that scales vision-centric post-training by using a set of lightweight puzzle environments, promoting exploration through curriculum design, and monitoring and improving reasoning--answer consistency during training. PuzzleCraft instantiates three tasks inspired by classic pretext objectives for visual pretraining: PatchFit~\cite{he2022masked, caron2021emerging}, Rotation~\cite{gidaris2018unsupervised}, and Jigsaw~\cite{noroozi2016unsupervised, misra2020self, wang2025jigsaw-r1, wu2025visual-jigsaw}. Our puzzle environments are fully unsupervised, require no external verifiers or teacher models, and provide automatic rewards at scale.

PuzzleCraft targets three bottlenecks in puzzle-based RLVR. \textbf{(1)} Reward sparsity: we use Jigsaw as a core environment with a graded, partial-credit reward (fraction of correctly placed tiles), which rewards intermediate progress and penalizes localized errors. \textbf{(2)} Flat rewards and vanishing advantages: we introduce a curriculum that accounts for both difficulty and exploration. The difficulty term upweights medium-hard prompts using reward variance, while the exploration term downweights prompt groups whose rollouts collapse to the same solution. For binary puzzles, exploration is measured by class entropy over selected options; for Jigsaw, it is measured by diversity in the induced permutations, reflecting its combinatorial structure. \textbf{(3)} Consistency drift: we introduce a post-training diagnostic metric, Reasoning--Answer Consistency (RAC). We sample rollouts throughout training and measure whether the reasoning trace supports the final answer, then use RAC to study how our curriculum and consistency-aware rewards affect faithfulness over time.


Empirically, our exploration-aware curriculum consistently outperforms difficulty-only curricula. Ablations further show that both the curriculum and consistency-aware reward schemes~\cite{chen2025grpo-care} improve reasoning--answer consistency and translate to higher downstream accuracy. PuzzleCraft yields reliable gains across model families and evaluation settings. On Qwen2.5-VL-7B, our Jigsaw variant improves average performance across nine image benchmarks, outperforming the strongest puzzle-based baseline by 2.2 points. On Qwen2.5-VL-3B, our best variant improves over baselines by up to 1.39 points. The same trend holds for Qwen3-VL across scales: Jigsaw post-training increases Avg. over the corresponding Qwen3-VL-Instruct checkpoints by +2.6 (2B), +2.99 (4B), and +1.87 (8B). We also observe positive transfer to video reasoning benchmarks despite not using any video data during post-training.

\vspace{0.5em}
\noindent{\textbf{Contributions:}}
\begin{enumerate}[leftmargin=*,label=\textbf{--}]
\item \textbf{PuzzleCraft:} a supervision-free puzzle-based RLVR framework for scalable vision-centric post-training with a set of lightweight, verifiable environments.

\item \textbf{Exploration-aware curriculum:} dynamic weighting that accounts for both difficulty and rollout collapse, emphasizing medium-hard prompts while encouraging solution-space exploration, mitigating flat rewards and vanishing advantages.

\item \textbf{Consistency analysis:} we introduce RAC as a diagnostic for consistency drift, and show that our exploration-aware curriculum and consistency-aware reward schemes improve both faithfulness and downstream performance.
\end{enumerate}

%% file: sec/2_related_work.tex
\section{Related Work}
\label{sec:related_work}


\noindent\textbf{LLM/VLM RL post-training.}
RL post-training has played a central role in improving instruction-following and alignment in language models~\cite{ouyang2022traininglanguagemodelsfollow, rafailov2024directpreferenceoptimizationlanguage}. More recently, RL with verifiable rewards (RLVR) has emerged as a scalable way to induce stepwise reasoning by training on outcomes that can be checked automatically~\cite{guo2025deepseek}. Group-relative policy optimization (GRPO)~\cite{shao2024deepseekmath} and related objectives have been studied from multiple angles, including comparisons to alternative RL formulations~\cite{liu2025understandingr1zeroliketrainingcritical, zheng2025groupsequencepolicyoptimization, yu2025dapoopensourcellmreinforcement}, how RLVR interacts with supervised fine-tuning (SFT)~\cite{chen2025sft-or-rl, chu2025sftmemorizesrlgeneralizes}, and training efficiency and scaling behavior~\cite{cai2025trainingfreegrouprelativepolicy, lin2025cppoacceleratingtraininggroup}. Motivated by these results, recent work has adapted RLVR and GRPO-style post-training to vision--language models (VLMs)~\cite{shen2025vlm-r1, huang2025vision, wang2025vl, feng2025video, deng2025openvlthinker}. Many efforts target multimodal reasoning on math- and science-oriented benchmarks, while others extend RL post-training to more vision-centric settings such as grounding and segmentation~\cite{liu2025segzeroreasoningchainguidedsegmentation, bai2025univgr1reasoningguideduniversal}. Despite this progress, scaling RLVR in the visual domain remains difficult: reliable verification is often task-specific and many pipelines still depend on curated annotations or external verifiers, which limits throughput and portability across tasks.

\vspace{0.3em}
\noindent\textbf{Towards supervision-free post-training.}
Obtaining clean ground-truth annotations can be costly, impractical at scale, and noisy, motivating post-training methods that reduce or remove dependence on labeled answers. In LLMs, several directions have been explored, including confidence-based heuristics (e.g., entropy-based selection)~\cite{prabhudesai2025maximizingconfidenceimprovesreasoning}, majority voting and self-consistency across rollouts~\cite{chen2025selfquestioninglanguagemodels, zuo2025ttrltesttimereinforcementlearning}, and training strategies that tolerate imperfect reward signals~\cite{shao2025spuriousrewardsrethinkingtraining}. For VLMs, verifier-based pipelines~\cite{wang2025vicrit, wang2025llavacriticr1criticmodelsecretly} (e.g., critic/judge models that assess captions or textual responses) and gamified or self-play environments~\cite{wang2025vision-zero, chen2025learning-images} improve perception and reasoning but introduce new costs and biases through external evaluators. Closer to our setting, recent work introduces visual puzzle tasks for post-training VLMs~\cite{wu2025visual-jigsaw, wang2025jigsaw-r1, zeng2025agentic, feng2025visual-sphinx}. These studies, while promising, typically cover a narrow puzzle set, rely on vanilla GRPO, and offer limited analysis of training dynamics, difficulty, and generalization. PuzzleCraft instead focuses on these dynamics, introducing an exploration-aware curriculum and consistency monitoring to improve stability and transfer.


\vspace{0.3em}
\noindent\textbf{Curricula for GRPO post-training.}
A growing body of work examines failure modes of CoT-enabled VLMs under GRPO-style training~\cite{jiang2025mmecotbenchmarkingchainofthoughtlarge, liao2025long-perceptual}. From an optimization perspective, vanilla GRPO is largely difficulty-agnostic: when rollouts within a group become homogeneous, group-relative advantages shrink toward zero and updates become weak~\cite{wang2025anglesdontlieunlocking, guo2025observe-r1, wang2025dumpautomateddistributionlevelcurriculum}. Sparse rewards can further worsen this behavior by pushing groups toward all-success or all-failure regimes~\cite{xia2025visionary-r1, chen2025grpo-care}. Recent approaches mitigate these issues through offline or online curricula and by introducing more efficient or stabilized GRPO variants~\cite{jiang2025vcrlvariancebasedcurriculumreinforcement, hammoud2025trainlongthinkshort, zheng2025groupsequencepolicyoptimization}. However, these curricula typically rely on reward statistics alone and do not capture whether the policy is exploring diverse solutions or collapsing to a repeated guess. PuzzleCraft extends this line of work by measuring solution-space dispersion within each rollout group and using it to downweight collapsed groups.

\vspace{0.3em}
\noindent\textbf{Consistency drift in reasoning.}
A recently highlighted failure mode of CoT is reasoning--answer inconsistency, where the final answer is not supported by the reasoning trace~\cite{shen2025mitigating, yao2025reasoningmodelspronehallucination}. Prior work reports that faithfulness can improve early during GRPO-style training but later degrade, motivating closer monitoring of post-training dynamics~\cite{chen2025reasoningmodelsdontsay}. Complementary methods propose consistency-aware objectives or auxiliary checks that better couple the final answer to the reasoning chain~\cite{chen2025grpo-care, huang2025answer-consistent}. In our setting, we observe the same trend in visual puzzle post-training: vanilla GRPO exhibits worsening consistency over time. We therefore introduce a simple diagnostic metric tracked throughout training, and find that our exploration-aware curriculum mitigates the decline. When combined with a lightweight consistency-aware GRPO variant (GRPO-CARE~\cite{chen2025grpo-care}), the gains in both consistency and downstream performance are further amplified.

%% file: sec/3_method.tex
\section{PuzzleCraft}
\label{sec:method}

\begin{figure*}[t]
\centering
\includegraphics[clip, trim=10cm 3cm 0cm 0cm, width=1.0\linewidth]{assets/method_v22.pdf}
\vspace{-1em}
\caption{\textbf{PuzzleCraft overview.} We sample puzzle prompts and weight them by a curriculum. The model generates a group of candidate solutions, receives automatically verifiable rewards, and updates the policy using a curriculum-weighted GRPO objective. We also track reasoning--answer consistency during post-training and study how curricula and consistency-aware optimization affect both RAC and downstream performance.}
\label{fig:visual_puzzles}
\vspace{-1em}
\end{figure*}

This section presents PuzzleCraft, a supervision-free RLVR framework that leverages automatically verifiable puzzle rewards to scale vision-centric post-training. Recent work has shown that puzzles provide a practical source of automatically verifiable rewards, yet existing pipelines typically follow vanilla GRPO recipes and treat puzzles as purely outcome-driven tasks. Here, we reframe puzzle-based RLVR as a curriculum and consistent reasoning problem rather than a pure reward-maximization objective. Specifically, we consider multiple vision-centric puzzle environments and focus on two aspects that become central in this setting: (i) how to weight prompts so that GRPO updates concentrate on informative, medium-hard instances while avoiding rollout collapse, and (ii) how to monitor and improve reasoning--answer consistency, which can degrade during post-training even as puzzle rewards continue to rise.

Our approach has three components. First, we instantiate a set of lightweight puzzle environments with built-in verification. Second, we introduce a curriculum that is both \textbf{difficulty-aware} and \textbf{exploration-aware}: in addition to mean success, it detects when rollouts within a prompt group collapse to the same solution, and it downweights such groups to mitigate vanishing-advantage updates. Third, we track a simple Reasoning--Answer Consistency (RAC) metric during post-training, and optionally incorporate GRPO-CARE~\cite{chen2025grpo-care} as an add-on to explicitly shape consistency. \autoref{fig:visual_puzzles} provides an overview of our approach.

\subsection{Verifiable rewards via puzzle environments}
\label{subsec:puzzles}

Motivated by classic self-supervised pretext tasks for pretraining, we instantiate three programmatically verifiable puzzle environments (see \autoref{fig:visual_puzzles}): Rotation~\cite{gidaris2018unsupervised} (predict an angle from a fixed set), PatchFit (identify a masked patch among confusable candidates), and Jigsaw~\cite{noroozi2016unsupervised, misra2020self, wang2025jigsaw-r1, wu2025visual-jigsaw} (assign tiles to grid positions under a valid permutation). Rotation and PatchFit yield binary rewards, $r\in\{0,1\}$, via exact checks.

Jigsaw has a property that we explicitly leverage. It naturally supports partial credit enabling graded supervision without process-level labels. We use a reward, $r\in[0,1]$, equal to the fraction of tiles placed correctly. This rewards intermediate progress and penalizes localized errors without collapsing the entire rollout to failure.

Each puzzle exposes a controllable difficulty parameter: grid size for Jigsaw, angle-set cardinality for Rotation, and distractor hardness for PatchFit. In our experiments, we keep these settings fixed when studying curricula, and analyze puzzle choice and inter-puzzle transfer separately in \S\ref{sec:experiments}.


\subsection{Difficulty and exploration aware curriculum}
\label{subsec:curriculum}

For each prompt, $x$, we sample a group of $G$ rollouts, $\{o_i\}_{i=1}^{G}$, with rewards ,$\{r_i\}$. We assign the group a weight, $w(x)$, so that training focuses on prompts that yield informative GRPO updates.

\vspace{0.2em}
\noindent{\textbf{Why difficulty alone is insufficient.}}
Difficulty-aware curricula in RLVR often rely on reward statistics such as mean success or reward variance, which emphasize medium-difficulty prompts. This is necessary, but in puzzle RL it does not distinguish whether the model is exploring different candidate solutions or repeatedly collapsing to the same guess. As a simple example with a multiple-choice verifier and ground-truth option $A$ and incorrect options $B-F$, the groups $\{A,B,B,B,B,B,B,B\}$ and $\{A,B,B,C,C,D,E,F\}$ have the same mean success, yet the first reflects strong collapse to a single incorrect option. Collapsed groups are particularly problematic for GRPO because homogeneous rollouts cause group-relative advantages to shrink, even when the mean success suggests the prompt is nontrivial. This highlights that reward statistics alone are insufficient to characterize training signal quality in puzzle-based RLVR.

\vspace{0.2em}
\noindent{\textbf{Difficulty signal.}}
We use the group mean reward as the primary difficulty signal,
\[
d(x) \;=\; \bar r \;=\; \tfrac{1}{G}\sum_{i=1}^{G} r_i \in [0,1],
\]
where $d\approx 1$ indicates easy prompts and $d\approx 0$ indicates hard prompts. The term $d(1-d)$ peaks at medium difficulty and vanishes at the extremes, matching the regime where GRPO provides the most informative learning signal.

\vspace{0.2em}
\noindent{\textbf{Exploration signal.}}
To detect collapse, we measure the diversity in the puzzle solution space within the rollout group. The definition is puzzle-specific but follows a common principle: extract a discrete solution key, $s_i$, from each rollout and compute a normalized dispersion statistic, $d_{\pi}(x)\in[0,1]$.

\textbf{Binary puzzles (Rotation, PatchFit).}
Each rollout selects a discrete option (angle class for Rotation, candidate index for PatchFit). Let $s_i\in\{1,\dots,K\}$ denote the selected option and $\hat p_k=\tfrac{1}{G}\sum_{i=1}^{G}\mathbf{1}[s_i=k]$ be the empirical frequency. We define the normalized solution entropy as
\[
d_{\pi}(x) \;=\; \frac{-\sum_{k=1}^{K} \hat p_k \log \hat p_k}{\log K},
\]
where, $d_{\pi}\approx 0$ indicates that rollouts collapse to a single option, while larger values indicate broader exploration across options.

\textbf{Jigsaw.}
Jigsaw is a structured puzzle with partial grading, where we score each rollout by the fraction of correctly placed tiles. A useful side effect of this grading is that multiple \emph{distinct} permutations can yield the same reward, so reward statistics alone can miss whether the model is collapsing to a single arrangement. We therefore track exploration directly in permutation space. Each rollout induces a permutation over grid positions. Let $\Pi(o_i)\in\mathfrak{S}$ denote the induced permutation, and define
\[
M \;=\; \Bigl|\{\Pi(o_i)\}_{i=1}^{G}\Bigr|,\qquad 
d_{\pi}(x) \;=\; \frac{M-1}{G-1}.
\]
This quantity approaches $0$ when rollouts collapse to the same arrangement and increases as the group explores distinct permutations, including cases where rollouts achieve similar partial-credit rewards through different tile placements.

\vspace{0.2em}
\noindent{\textbf{Curriculum weight.}}
We combine difficulty and exploration into a single group weight:
\begin{equation}
\label{eq:curriculum}
w(x) \;=\; \lambda \, d(x)\,\bigl(1-d(x)\bigr)\,\bigl(d_{\pi}(x)\bigr)^{\gamma},
\end{equation}
where we use fixed values of $\lambda=4$ and $\gamma = 0.5$ in all experiments. The term $d(x)(1-d(x))$ emphasizes medium difficulty and becomes zero when all rollouts succeed or all fail. The term $d_{\pi}(x)$ downweights groups whose rollouts collapse to the same solution, which are most prone to weak or uninformative group-relative updates. For Rotation and PatchFit, $d_{\pi}$ is computed from class entropy; for Jigsaw, it is computed from permutation diversity.

\vspace{0.2em}
\noindent{\textbf{Optimization objective.}}
We adopt a token-level GRPO objective with curriculum weighting. We set the KL-to-reference coefficient to zero ($\beta=0$), following recent guidance~\cite{chen2025grpo-care, guo2025observe-r1} that strong KL anchoring can over-constrain exploration in GRPO-style post-training:
\begin{align}
\label{eq:pcgrpo}
\mathcal{J}_{\text{PuzzleCraft}}(\theta)
&= \mathbb{E}_{(q,a)\sim\mathcal{D},\, \{o_i\}_{i=1}^{G}\sim \pi_{\theta_{\mathrm{old}}}(\cdot \mid q)}\!\\
&\hspace{-5em}\Bigg[
\tfrac{1}{G}\!\sum_{i=1}^{G}\underbrace{w(d(q))}_{\text{curriculum}}\;\tfrac{1}{|o_i|}\!\sum_{t=1}^{|o_i|}
\min\!\Big(\rho_{i,t}\,\hat A_{i,t},\,\tilde\rho_{i,t}\,\hat A_{i,t}\Big)\Bigg], \nonumber
\end{align}
where given $\epsilon > 0$:
\[
A_i = r_i-\bar r,\qquad \hat A_{i,t}\;=\;A_i,
\]
\[
\rho_{i,t}=\frac{\pi_{\theta}\!\big(o_{i,t}\mid q, o_{i,<t}\big)}{\pi_{\theta_{\mathrm{old}}}\!\big(o_{i,t}\mid q, o_{i,<t}\big)},\;\;\;\;
\tilde\rho_{i,t}=\mathrm{clip}\!\big(\rho_{i,t},\,1{-}\epsilon,\,1{+}\epsilon\big).
\]

\subsection{Consistency monitoring and consistency-aware optimization}
\label{subsec:rac}

Recent work highlighted that RL post-training can introduce a gap between the reasoning trace and the final answer, and that this gap may widen over training~\cite{chen2025reasoningmodelsdontsay,chen2025grpo-care}. For puzzle RL, this is easy to miss if one only monitors puzzle reward, since reward can improve while faithfulness degrades. We therefore monitor Reasoning--Answer Consistency (RAC) as a training-dynamics signal.

\vspace{0.2em}
\noindent{\textbf{RAC metric (diagnostic only).}}
To monitor reasoning--answer alignment during post-training, we periodically sample rollouts and compute RAC as a \emph{diagnostic} signal. Specifically, we prompt a fixed open-source judge (Qwen2.5-VL-72B in inference mode) with each rollout's rationale and final \texttt{<answer>}, and ask whether the rationale explicitly supports the emitted answer. Each trial is scored in $[0,1]$ and we report a moving average over training steps. Importantly, RAC is \emph{not} used for reward shaping, sample reweighting, or gradient updates in our training pipeline. Empirically, RAC tends to increase early in training and may drift downward later, consistent with observations in GRPO-style post-training. In \S\ref{supp::sec:rac_exp_setup} we provide the full evaluation setup, including the exact prompts and scoring procedure used to compute RAC.

\vspace{0.2em}
\noindent{\textbf{Consistency-aware add-on.}}
Consistency shaping is complementary to curriculum design, so we optionally incorporate GRPO-CARE~\cite{chen2025grpo-care} as an add-on to PuzzleCraft. Empirically, our curriculum mitigates late-stage RAC degradation, and adding CARE further improves both RAC and downstream performance. We use RAC as a diagnostic throughout and study its interaction with reward design and curricula in \S\ref{sec:experiments}.

%% file: sec/4_exps.tex
\section{Experiments}
\label{sec:experiments}

We first describe our implementation details and evaluation setup. We then run ablations to study the main design choices in PuzzleCraft. Finally, using the best-performing configuration, we train models across different scales and backbones and evaluate them on a broad suite of vision-centric benchmarks.

\subsection{Experimental Setup}
\noindent\textbf{Training data.}
We train on the COCO 2014~\cite{lin2014microsoft} training split (82{,}783 images). We choose COCO for reproducibility and to avoid introducing new images beyond those already seen during base-model pretraining, which helps isolate the effect of GRPO in our setting. For each image, we synthesize a single puzzle instance (Jigsaw, Rotation, or PatchFit). Unless stated otherwise, each model is trained on 82{,}783 puzzle prompts. We apply no additional preprocessing or filtering. Examples are shown in Fig.~\ref{fig:visual_puzzles} (top), and full puzzle and prompt details are provided in \S\ref{supp:discussion_details}.

\vspace{0.3em}
\noindent\textbf{Training framework and models.}
We initialize from Qwen2.5-VL-Instruct and Qwen3-VL-Instruct checkpoints~\cite{bai2025qwen25vltechnicalreport}. Unless stated otherwise, ablations are conducted with Qwen2.5-VL-7B-Instruct. We build on the public VLM-R1 training code for GRPO post-training. When enabling the consistency-aware variant GRPO-CARE, we use the authors’ released implementation and match their hyperparameters where possible. All training runs are performed on 8$\times$A100 (80\,GB) GPUs.

\vspace{0.4em}
\noindent\textbf{Hyperparameters.}
Unless noted, we follow \texttt{VLM\mbox{-}R1} defaults with two changes: KL coefficient $\beta{=}0$ and learning rate $5{\times}10^{-7}$. We use batch size 16 for \texttt{VLM\mbox{-}R1} and batch size 8 for \texttt{GRPO\mbox{-}CARE}. All runs are trained for 1 epoch with maximum decoding length 2048. Each prompt uses $G{=}8$ rollouts with temperature $0.9$ and top-$p$ $0.95$, using one GRPO iteration per update. We train in bfloat16 and cap vision-encoder tokens at 1024 during post-training. We set the PPO clipping parameter $\epsilon{=}0.2$ for \texttt{VLM\mbox{-}R1}. When using \texttt{GRPO\mbox{-}CARE}, we adopt the authors’ defaults (\texttt{ref\_ema\_decay} $0.995$, EMA update every 10 steps, bonus coefficient $0.5$, confidence upper bound $0.95$, consistency margin $0.01$) and set $\epsilon{=}0$.

\vspace{0.4em}
\noindent\textbf{Evaluation setup.}
We evaluate primarily with \texttt{VLMEvalKit}~\cite{duan2024vlmevalkit}. Unless otherwise noted, all results are reported in \emph{thinking mode}: we append a standardized prompt to elicit \texttt{think}$\rightarrow$\texttt{answer} formatting, and use \texttt{Qwen-VL-2.5-72B} within \texttt{VLMEvalKit} for post-processing and format checking. The base prompt is:
\begin{quote}\small
\texttt{First output the thinking process in <think></think> tags and then\\ output the final answer in <answer></answer> tags.}
\end{quote}

Qwen3-VL-Instruct models do not support \texttt{<think></think>}, so we instead use \texttt{<analyze></analyze>} for the reasoning trace.

\vspace{0.3em}
\noindent\textbf{Baselines.}
In addition to the underlying base models, we compare against a broad set of recent RLVR frameworks with publicly released checkpoints. Our main points of comparison are puzzle-based methods, including ViCrit~\cite{wang2025vicrit}, Vision-Zero~\cite{wang2025vision-zero}, Visual Jigsaw~\cite{wu2025visual-jigsaw}, Jigsaw-R1~\cite{wang2025jigsaw-r1}, VisualSphinx~\cite{feng2025visual-sphinx}, and Game-RL~\cite{tong2025game}. We also report results against annotated RLVR approaches such as Vision-R1~\cite{huang2025vision}, VL-Rethinker~\cite{wang2025vl}, Video-R1~\cite{feng2025video}, ViGoRL~\cite{sarch2025grounded}, GRPO-CARE \cite{chen2025grpo-care}, and VLM-R1~\cite{shen2025vlm-r1}.

\vspace{0.3em}
\noindent\textbf{Image benchmarks.}
We evaluate on a wide range of multimodal benchmarks. For images, our suite includes math-oriented reasoning tasks (MathVista~\cite{lu2023mathvista}, MathVision~\cite{wang2024measuring}, MathVerse~\cite{zhang2024mathverse}) and vision-centric VQA benchmarks (MME \cite{fu2025mme}, MMStar~\cite{chen2024we}, POPE~\cite{li2023pope}, MMT-Bench~\cite{mmtbench}, CVBench-2D~\cite{cvbench}, and MMVP~\cite{tong2024mmvp}).

\vspace{0.2em}
\noindent\textbf{Video benchmarks.}
For video understanding and reasoning, we evaluate on VideoMME~\cite{fu2025video}, TempCompass~\cite{liu2024tempcompass}, Video-TT~\cite{zhang2025towards}, MVBench~\cite{li2024mvbench}, Q-Bench-Video~\cite{zhang2025q}, and Video-MMMU~\cite{hu2025video}, and CG-Bench~\cite{chen2024cg}.

\input{assets/tables/ablate_design}

\input{assets/figure4/figure4}

\subsection{Ablating design choices}
We ablate the main components of PuzzleCraft to understand their individual contributions and identify the most effective configuration for puzzle-based RLVR. Once this setup is established, we apply it across different model families and puzzle environments. All the ablations are done with Qwen2.5-VL-7B and on the Jigsaw puzzle environment on six vision-centric image benchmarks.

\vspace{0.3em}
\noindent\textbf{Reasoning--Answer Consistency.}
We study how reasoning--answer consistency evolves during post-training using the RAC metric from \S\ref{subsec:rac}. We compare four variants: vanilla GRPO, GRPO with our curriculum, GRPO-CARE, and curriculum+GRPO-CARE. Fig.~\ref{fig:scaling_4panels} tracks RAC alongside puzzle reward, reward variance, and response length over training.

Vanilla GRPO shows the clearest consistency drift: RAC increases early, then degrades later even as puzzle reward continues to improve, consistent with observations in prior work on GRPO-style post-training~\cite{chen2025reasoningmodelsdontsay}. Adding our curriculum mitigates the late-stage decline, while GRPO-CARE further improves RAC. The combined setup maintains the highest RAC through most of training, suggesting that curriculum weighting and consistency-aware optimization are complementary in puzzle-based RLVR.

\autoref{tab:ablate_design} reports downstream performance across six vision-centric benchmarks. Both the curriculum and consistency-aware optimization improve average accuracy. Notably, curriculum+GRPO-CARE outperforms GRPO-CARE alone, despite GRPO-CARE achieving higher puzzle reward during post-training. This highlights that higher puzzle reward is not, by itself, a reliable indicator of downstream reasoning performance.



\vspace{0.5em}
\noindent\textbf{Curricula.}
Most puzzle-based RLVR frameworks treat rewards as flat across difficulty levels. As noted in prior work, this can trigger vanishing-advantage dynamics: rollouts within a group become homogeneous, the group-relative signal collapses, and learning slows or stalls, wasting compute. A common remedy is difficulty-aware reweighting that emphasizes medium-difficulty samples during post-training. However, as discussed in \S\ref{subsec:curriculum}, these curricula typically rely only on reward statistics and do not reflect whether rollouts are actively exploring the solution space or have collapsed to a repeated guess. PuzzleCraft instead uses an exploration-aware curriculum that couples sample difficulty with a solution-space dispersion signal.


We compare our curriculum against a representative difficulty-aware baseline, Observe-R1~\cite{guo2025observe-r1}, with results reported in \S\ref{supp:sec:curricula}. On Jigsaw, the exploration-aware curriculum achieves the strongest downstream performance and outperforms Observe-R1 under the same training setup. This suggests that curricula which explicitly track rollout exploration, rather than reward statistics alone, can yield more effective puzzle-based post-training.

\subsection{Main Results}
\label{subsec:main_results}
\input{assets/tables/table_main}
\input{assets/tables/video_results}

Guided by the analysis above, we apply our best-performing configuration (curriculum+GRPO-CARE) to Rotation and PatchFit in addition to Jigsaw. Since different puzzles emphasize different skills, we also train a mixed setting to study multi-puzzle post-training. Concretely, we sample 40K training instances (15K Jigsaw, 15K PatchFit, 10K Rotation) and run post-training.

\autoref{tab:main_table} reports results on nine image benchmarks spanning math reasoning and vision-centric VQA for Qwen2.5-VL and Qwen3-VL across multiple model sizes. \autoref{tab:video_benchmarks} further compares our models against baselines on 7 video reasoning benchmarks.

\vspace{0.5em}
\noindent\textbf{Findings.}
\autoref{tab:main_table} and \autoref{tab:video_benchmarks} highlight the following trends:
\begin{itemize}[leftmargin=*,labelindent=0pt,label=\textbf{--}]
    \item \textbf{PuzzleCraft improves puzzle-based RLVR.}
    PuzzleCraft models, especially Jigsaw and Mix, achieve the strongest results among puzzle-based approaches. On image benchmarks, Qwen2.5-VL-7B-Jigsaw reaches 65.18\% Avg., improving by more than 2.2\% over the closest puzzle baseline (Visual Jigsaw). Similar gains hold for Qwen2.5-VL-3B and across Qwen3-VL sizes. On video benchmarks, Qwen2.5-VL-7B-Jigsaw also outperforms the strongest puzzle baseline (VisualSphinx) by about 1.5\%. These results suggest that puzzle-based RLVR still has meaningful headroom when paired with curricula that avoid rollout collapse and consistency-aware optimization.

    \item \textbf{Competitive with curated RLVR baselines.}
    Despite using only supervision-free puzzle training, our models are competitive with methods trained using large-scale curated RLVR data. For Qwen2.5-VL-7B, Jigsaw improves over GRPO-CARE by about 2\% on Avg. and is comparable to Vision-R1 and VL-Rethinker overall, trailing mainly on math benchmarks while improving on vision-centric ones. For Qwen2.5-VL-3B, Mix outperforms curated RLVR baselines (VLM-R1 and ViGoRL) by more than 2.5\%. On video benchmarks, Qwen2.5-VL-7B-Jigsaw is close to Video-R1 despite not using any video data during post-training.

    \item \textbf{Puzzle choice and transfer matter.}
    Different puzzles transfer differently to downstream tasks. Jigsaw performs best overall, which we attribute in part to its graded, partial-credit reward. Rotation also transfers well, but its gains are benchmark-dependent and complementary to Jigsaw. PatchFit transfers weakly on both math and vision-centric benchmarks. Mix is a robust choice across tasks..
\end{itemize}

\subsection{Direct Inference (Non-Thinking)}
All experiments in \S\ref{subsec:main_results} are reported under a CoT-inducing prompt. Here we also evaluate in direct inference mode, using default benchmark prompts that do not request an explicit reasoning trace. We focus on the Qwen2.5-VL-7B backbone and compare against representative baselines. Results are shown in \autoref{tab:7b_direct_mode}.

Consistent with prior observations on Qwen backbones and RLVR-tuned variants, some benchmarks achieve higher accuracy in direct mode than with CoT. Understanding this discrepancy remains an active research topic. Overall, our models are competitive with or outperform baselines across vision-centric tasks. At the same time, direct-mode results are generally close to the base model, suggesting that most RL gains in our setting emerge when the model is prompted to produce an explicit reasoning trace.

\input{assets/tables/table_7b_direct}

\subsection{Performance on Puzzles}
Motivated by the reported weakness of LLMs/VLMs on puzzle environments~\cite{lyu2025jigsaw, shojaee2025illusionthinkingunderstandingstrengths}, we evaluate how well our puzzle training improves model performance and whether the learned skills transfer across puzzle types. To create the test set, we randomly select 1000 samples from the test images of COCO2014 and create Jigsaw, PatchFit, and Rotation puzzles. As shown in \autoref{tab:puzzles}, we clearly observe the challenges of reasoning models with our setup as well. Specially, we can see that post-training on a puzzle environment improves performance on that one, but the gains do not transfer to other puzzle setups, but the performance even degrades compared to the Qwen baseline, a mixture of environments alleviates this problem, however, there is still no reliable way to indicate whether learning one skills transfers to others.

\input{assets/tables/table_puzzles}

%% file: assets/tables/ablate_design.tex
\begin{table*}[t]
\centering
\footnotesize
\setlength{\tabcolsep}{4pt}
\renewcommand{\arraystretch}{1}

\caption{Ablations of PuzzleCraft design choices on vision-centric benchmarks.  consistency variants (CL (curriculum), CARE, and their combination). Avg. denotes the mean across benchmarks; MME is normalized by dividing by 2800.}
\vspace{-0.5em}
\begin{adjustbox}{max width=\textwidth}
    
		\begin{tabular}{l|c c c c c c  | c}
			\hline
			\textbf{Design choice} & MME &	MMStar &	POPE &	MMT &	CV-Bench& MMVP   & Avg.\\
			\hline
			 
			Qwen2.5-VL-Instruct & 2243 & 64.67 & 81.77 & 59.64 & 73.62 & 77.67 & 72.91 \\ \hline

            \rowcolor[HTML]{EFEFEF} \multicolumn{8}{l}{\textbf{Consistency}} \\\hline
            
            Jigsaw  & 2340 & 64.47 & 85.17 & 59.96 & 72.13 & 75.33 & 73.44 \\
			Jigsaw w/ CARE & 2319 & 65.80 & 86.95 & 61.18 & 77.76 & 77.00 & 75.25 \\
			Jigsaw w/ CL  & 2365 & 64.27 & 84.35 & 62.62 & 75.37 & 74.67 & 74.30 \\
			Jigsaw w/ CL + CARE (Main) & 2366 & 64.60 & 86.52 & 62.26 & 77.63 & 78.67 & 75.70 \\

			\hline
		\end{tabular}%
    \end{adjustbox}

 	\label{tab:ablate_design}


\end{table*}

%% file: assets/figure4/figure4.tex
\begin{figure*}[t]
  \centering
  \begin{subfigure}[t]{0.48\textwidth}
    \centering
    \includegraphics[width=\linewidth]{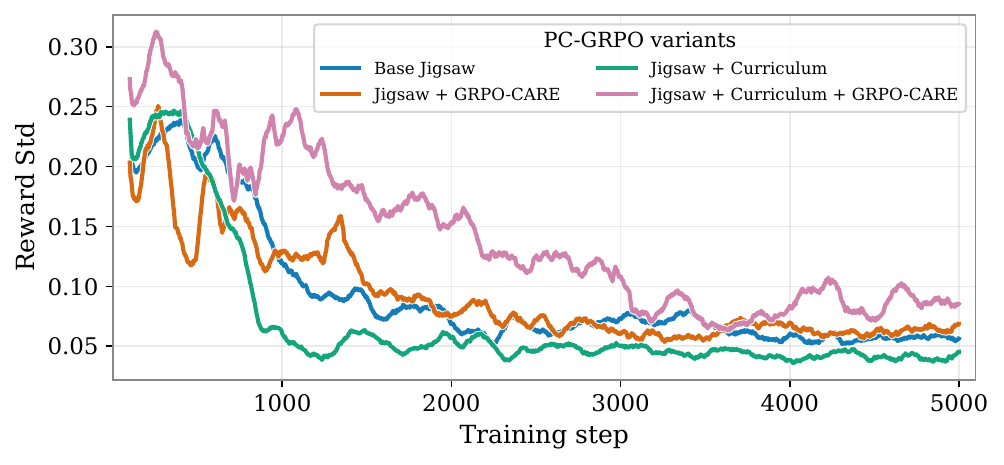}
    \subcaption{Reward variance}
    \label{fig:perplexity_graph}
  \end{subfigure}
  \hfill
  \begin{subfigure}[t]{0.48\textwidth}
    \centering
    \includegraphics[width=\linewidth]{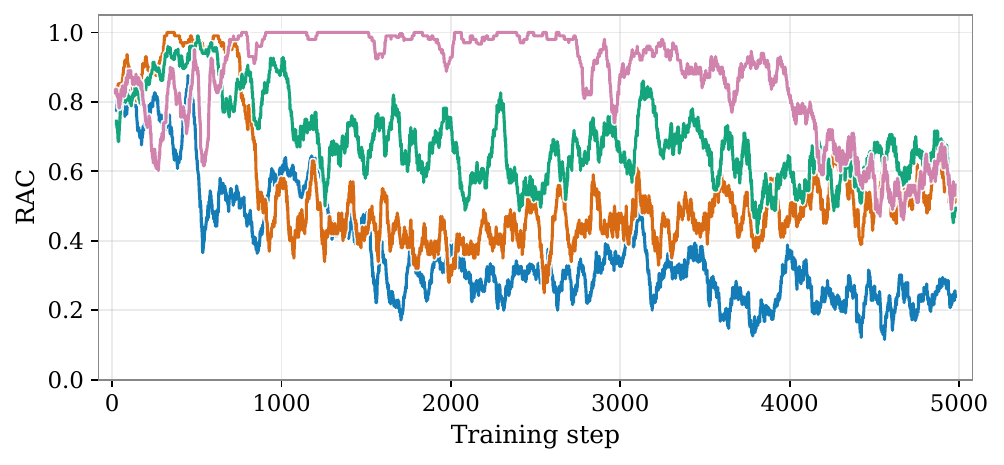}
    \subcaption{Reasoning-answer consistency}
    \label{fig:reasoning_graph}
  \end{subfigure}

  \vspace{0.4em} 

  \begin{subfigure}[t]{0.48\textwidth}
    \centering
    \includegraphics[width=\linewidth]{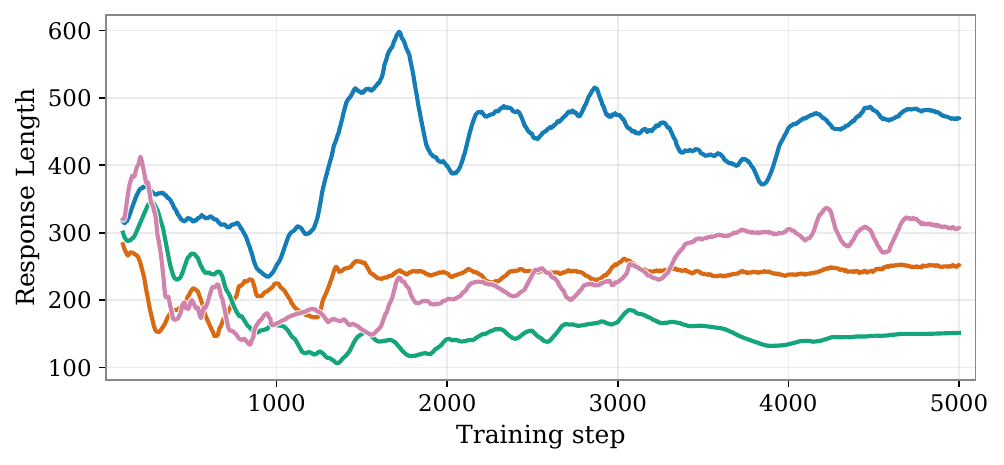}
    \subcaption{Response length (in tokens)}
    \label{fig:vision_graph}
  \end{subfigure}
  \hfill
  \begin{subfigure}[t]{0.48\textwidth}
    \centering
    \includegraphics[width=\linewidth]{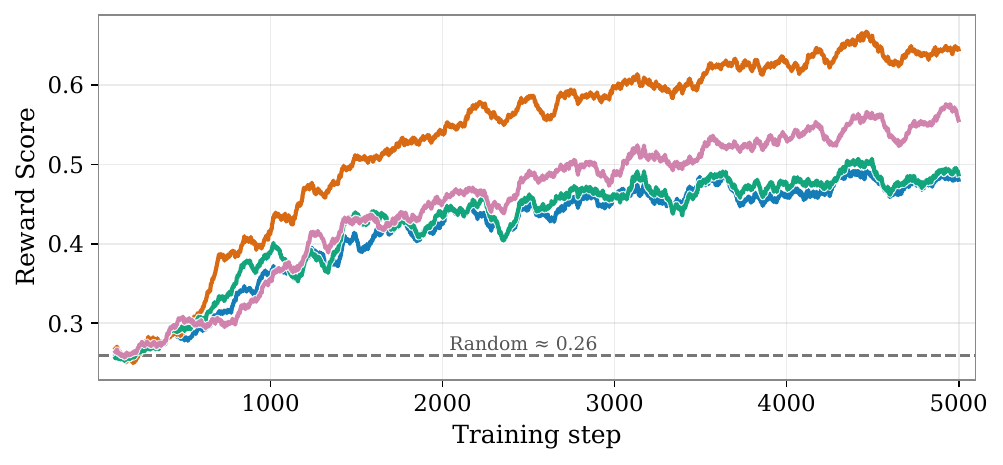}
    \subcaption{Reward score}
    \label{fig:ablations_graph}
  \end{subfigure}

  \caption{Tracking GRPO metrics during post-training across four puzzle environments. All charts report a moving average with window size of 100 over training steps.
    \textbf{(a)} Variance among the rollout rewards
    \textbf{(b)} Consistency rate between rollout reasoning and final answer, measured by Qwen2.5-VL-72B model
    \textbf{(c)} Average numbers of tokens decoded by each trajectory
    \textbf{(d)} Reward score which is the partially graded Jigsaw solution reward.}
  \label{fig:scaling_4panels}
\end{figure*}

%% file: assets/tables/table_main.tex
\renewcommand{\arraystretch}{1.2}
\begin{table*}[!t]

    \caption{Results on image reasoning benchmarks for Qwen2.5-VL and Qwen3-VL backbones. We compare supervised RLVR baselines, puzzle-based RLVR baselines, and PuzzleCraft variants (single-puzzle and Mix). Avg. is the mean over all 9 benchmarks, with MME normalized as $(\mathrm{MME}/2800)\times 100$ before averaging. PuzzleCraft delivers consistent gains across model sizes and backbones, with Jigsaw and Mix outperforming prior puzzle-based RLVR and remaining competitive with supervised RLVR baselines.}
 	\label{tab:main_table}
    \vspace{-0.5em}

	\resizebox{\textwidth}{!}{%
		\begin{tabular}{l| c c c | c c c c c c  | c}
			\hline
            \multirow{2}{*}{\textbf{Model}}
            & \multicolumn{3}{c|}{\textbf{Math}}
            & \multicolumn{6}{c}{\textbf{Image VQA}}
            &
            \\
            
            \cmidrule(lr){2-4}
            \cmidrule(lr){5-11}

			  & MathVista & MathVision & MathVerse   & MME &	MMStar &POPE &	MMT &	CV-Bench & MMVP   &  Avg.\\
			\hline
			 \rowcolor[HTML]{EFEFEF} \multicolumn{11}{l}{\textbf{Qwen2.5-VL-7B}} \\\hline
             \rowcolor[HTML]{DDEFFF} \multicolumn{11}{c}{\textbf{Baselines}} \\
            \hline
			Qwen 2.5 VL (Vanilla) & 66.30 & 23.68 & 56.49 &  2243 & 64.67 & 81.77 & 59.64 & 73.62 & 77.67 & 64.88 \\
            Vision-R1 (ICLR26) & 68.89 & 39.47 & 59.99 &  2292 & 62.33 & 88.41 & 59.70 & 72.17 & 77.00 & 67.76 \\
            VL-Rethinker (NeurIPS25) & 72.39 & 29.90 & 67.29 &  2311 & 63.06 & 83.62 & 61.01 & 76.90 & 77.33 & 68.23 \\
            GRPO-CARE (25.06) & 68.70& 20.39& 47.71&  2352 & 64.13 & 88.18 & 62.62 & 74.91 & 80.33 & 65.66 \\
            \hline
            
            \hline
			ViCrit (NeurIPS25) & 61.40& 18.75 & 38.02&  2167 & 62.27 & 80.50 & 59.83 & 70.84 & 75.33 & 60.48 \\
			Vision-Zero (ICLR26) & 66.20 &21.05&45.86 &  2248 & 63.47 & 82.67 & 60.47 & 73.53 & 78.00 & 63.50 \\
			Visual Jigsaw (ICLR26) & 67.50 & 29.27 & 57.50 &  2243 & 62.53 & 84.78 & 57.76 & 74.46 & 76.33 & 65.58 \\
			VisualSphinx (25.05) & 67.80 &26.31 &53.90 &  2296 & 63.20 & 83.93 & 60.63 & 73.98 & 77.33 & 65.45 \\
            Game-RL (ICLR26) & 67.40 & 24.01 & 58.00 & 2229 & 64.60 & 81.77 & 61.27 & 75.24 &77.66 &  65.51 \\
			 \hline
			
			\rowcolor[HTML]{DDEFFF} \multicolumn{11}{c}{\textbf{Our Variants}} \\
			\hline
			Jigsaw  & 68.20 & 30.92 & 56.70 &  2366 & 64.60 & 86.52 & 62.26 & 77.63 & 78.67 & \textbf{67.78} \\
			PatchFit   & 68.03 & 26.31 & 50.89 &  2316 & 59.87 & 85.05 & 58.94 & 73.37 & 78.00 & 64.80  \\
			Rotation  & 71.70 & 22.69 & 57.70 &  2357 & 64.60 & 87.36 & 61.91 & 75.08 & 79.67 & 67.21 \\
			Mix  & 68.20 & 24.01& 58.10 &  2359 & 65.20 & 85.40 & 62.65 & 76.96 & 78.33 & 67.01  \\
			\hline
			\hline
            \rowcolor[HTML]{EFEFEF} \multicolumn{11}{l}{\textbf{Qwen2.5-VL-3B}} \\\hline

            \rowcolor[HTML]{DDEFFF} \multicolumn{11}{c}{\textbf{Baselines}} \\
            \hline
            Qwen 2.5 VL (Vanilla)  & 57.19 & 21.38 &  38.30 &  2180 & 54.73 & 77.41 & 53.31 & 65.62 & 63.33 & 56.57 \\
            VLM-R1 (25.04)  & 57.99 & 19.73 & 40.80 &  2207 & 55.20 & 79.66 & 52.06 & 66.65 & 69.67 & 57.84 \\
            ViGoRL (NeurIPS25) & 56.10 & 18.42 & 34.60 &  1919 & 50.46 & 84.75 & 54.90 & 79.21 & 68.66 & 57.29 \\
            \hline
            Jigsaw-R1 (TMLR25)  & 58.80 & 22.69 & 40.30 &  2184 & 55.53 & 78.05 & 57.53 & 70.87 & 69.66 &  59.05\\
            \hline

            \rowcolor[HTML]{DDEFFF} \multicolumn{11}{c}{\textbf{Our Variants}} \\
            \hline

            Jigsaw  & 58.09 & 18.42 & 44.70 &  2223 & 55.40 & 78.68 & 57.88 & 71.33 & 68.67 & 59.17 \\
            Mix  & 60.60 & 24.01& 49.00 &  2127 & 57.53 & 77.30 & 57.72 & 72.88 & 69.0 & \textbf{60.44} \\
            
            \hline
			\hline
            \rowcolor[HTML]{EFEFEF} \multicolumn{11}{l}{\textbf{Qwen3-VL}} \\\hline

            Qwen3-VL-2B-Instruct  & 40.69 & 15.13& 28.10&  2072 & 46.6 & 82.88 & 53.53 & 71.36 & 70.33 & 53.62 \\
            Qwen3-VL-2B-Jigsaw (Ours)  & 42.00 & 19.07& 37.50&  2076 & 48.86 & 84.81 & 56.53 & 72.10 & 71.00 & \textbf{56.22} \\
            \hline

            Qwen3-VL-4B-Instruct  & 51.63 & 22.03 & 44.30&  2138 & 53.40 & 87.57 & 58.77 & 73.98 & 75.66 & 60.41 \\
            Qwen3-VL-4B-Jigsaw (Ours)  & 53.40 & 28.28 & 51.90&  2207 & 55.80 & 87.56 & 61.24 & 76.55 & 76.33 &  \textbf{63.32} \\
            \hline

            Qwen3-VL-8B-Instruct  & 57.40& 23.68 & 47.90&  2243 & 57.46 & 86.25 & 59.96 & 76.84 & 75.33 & 62.77  \\
            Qwen3-VL-8B-Jigsaw (Ours)  & 58.19 & 28.28& 55.20&  2227 & 57.66 & 86.58 & 61.72 & 76.56 & 78.00 & \textbf{64.64}  \\

            \bottomrule

		\end{tabular}
	}

 \vspace{-2em}
\end{table*}

%% file: assets/tables/video_results.tex
\renewcommand{\arraystretch}{1.2}
\begin{table*}[t]
\centering
\footnotesize
\setlength{\tabcolsep}{4pt}
\renewcommand{\arraystretch}{1}
\caption{Results on the seven video reasoning benchmarks. We compare PuzzleCraft models against recent RLVR baseline; Avg. denotes the mean across video benchmarks. PuzzleCraft gains transfer to video reasoning without video post-training, surpassing puzzle-based baselines and approaching video-supervised methods on several benchmarks.}
\label{tab:video_benchmarks}
\vspace{-0.4em}

\begin{adjustbox}{max width=\textwidth}

    \begin{tabular}{l|c c c c c c c | c}
        \hline
        \textbf{Method} & CGBench & Video-MMMU & MVBench & TempCompass & Video-MME & Video-TT & QBench-Video & Avg. \\
        \hline

        Qwen2.5-VL-Instruct & 26.71 & 33.83 & 55.80 & 55.00 & 65.40 & 34.40 & 58.50 & 47.09 \\ \hline

        \rowcolor[HTML]{EFEFEF} \multicolumn{9}{l}{\textbf{Baselines}} \\\hline

        VisualJigsaw          & 31.56 & 32.33 & 55.60 & 66.26 & 66.20 & 37.40 & 57.48 & 49.54 \\
        VisualSphinx          & 31.17 & 43.16 & 57.40 & 70.00 & 72.00 & 36.40 & 57.14 & 52.46 \\
        Video-R1 (NeurIPS25)             & 37.00 & 40.16 & 63.40 & 78.20 & 72.20 & 39.40 & 55.44 & 55.11 \\
        ViCrit          & 35.85 & 34.00 & 58.40 & 54.40 & 72.20 & 36.60 & 56.80 & 49.75 \\
        GRPO-CARE          & 33.94 & 34.50 & 62.00 & 67.60 & 62.80 & 38.80 & 54.76 & 50.62 \\
        Vision-Zero             & 27.48 & 37.50 & 60.00 & 51.20 & 67.80 & 34.40 & 58.16 & 48.07 \\
        \hline

        \rowcolor[HTML]{DDEFFF} \multicolumn{9}{l}{\textbf{Our variants}} \\
        \hline

        Jigsaw          & 36.33 & 38.00 & 59.00 & 75.40 & 71.40 & 38.60 & 58.84 & \textbf{53.93} \\
        Mix             & 34.86 & 42.66 & 61.00 & 71.80 & 71.00 & 34.59 & 56.12 & 53.14 \\

        \hline
    \end{tabular}%
\end{adjustbox}


\end{table*}

%% file: assets/tables/table_7b_direct.tex
\renewcommand{\arraystretch}{1}
\begin{table*}[!t]
\caption{Performance of 7B models on vision-centric benchmarks in direct-mode prompting (no explicit CoT). Avg. is the mean across benchmarks. Although direct-mode results are generally close to the base model, our Mix setup achieves the best overall Avg., suggesting that most RLVR gains emerge when models are prompted to produce an explicit reasoning trace.}
 	\label{tab:7b_direct_mode}
    \vspace{-0.5em}
	\resizebox{\textwidth}{!}{%
		\begin{tabular}{l|c c c c c c | c}
			\hline
			Model & MME&	MMStar & POPE &	MMT &CV-Bench & MMVP  & Avg.\\
			\hline
            Qwen2.5-VL-Instruct & 2308 & 63.27 & 86.37 & 62.74 & 76.29 & 78.33  & 74.90 \\ \hline
			\rowcolor[HTML]{EFEFEF} \multicolumn{8}{l}{\textbf{Baselines}} \\\hline
			ViCrit & 2178 & 64.13 & 85.96 & 62.58 & 76.59 & 76.33 & 73.90 \\
			Vision-Zero & 2306 & 63.53 & 85.92 & 63.06 & 76.00 & 77.67  & 74.76  \\
			Visual Jigsaw & 2313  & 64.13 & 86.37 & 62.20 & 76.47 & 79.00  & 75.13 \\
			VisualSphinx & 2341 & 63.73 & 86.34 & 62.65 & 76.76 & 77.33  & 75.07  \\
			GRPO-CARE & 2355 & 63.93 & 86.85 & 63.58 & 75.81 & 78.00  & 75.38 \\ \hline

			\rowcolor[HTML]{DDEFFF} \multicolumn{8}{l}{\textbf{Our variants}} \\
			\hline
			Jigsaw & 2348 & 64.33 & 86.05 & 63.70 & 76.12 & 77.33  & 75.23 \\
			Mix & 2371 & 64.73 & 86.65 & 64.09 & 77.36 & 78.67  & \textbf{76.03} \\
			\hline
		\end{tabular}%
	}

\end{table*}

%% file: assets/tables/table_puzzles.tex
\renewcommand{\arraystretch}{1}
\vspace{-1em}
\begin{table}[]
\centering
\setlength{\tabcolsep}{3pt}

\caption{Evaluating inter-puzzle transferability across our three puzzle environments. Training on a specific puzzle consistently improves performance, but gains do not transfer to others. Mixed-puzzle training yields strong gains across all tasks.}
\label{tab:puzzles}
\vspace{-0.5em}

\begin{tabular}{l|c c c c c}
\hline
\textbf{Model}& Jigsaw  & PatchFit & Rotation \\ \hline
Qwen2.5-VL-Instruct & 25.59 & 21.2 & 53.3\\
\hline
Jigsaw & \underline{36.65} & 21.7 & 33.9 \\ 
PatchFit  & 17.88 & \textbf{62.6} & 51.8  \\

Rotation & 25.55 & 20.9  & \underline{70.8} \\
Mix & \textbf{36.83} & \underline{48.6} & \textbf{83.2}  \\ 
\hline
\end{tabular}
\vspace{-1em}

\end{table}

%% file: sec/5_conclusion.tex

\section{Conclusion}
\label{sec:conclusion}

We studied puzzle environments as a scalable source of supervision-free RL with verifiable rewards (RLVR) for vision--language model post-training. We introduced \textbf{PuzzleCraft}, which combines lightweight, automatically verifiable puzzles with an exploration-aware curriculum that accounts for both difficulty and rollout collapse, and a simple Reasoning--Answer Consistency (RAC) diagnostic with optional consistency-aware optimization. Across Qwen2.5-VL and Qwen3-VL backbones, PuzzleCraft improves robustness and consistency on a broad suite of image benchmarks, and transfers to video reasoning without using video data during post-training.

Our results suggest that progress in puzzle-based RLVR is limited less by the availability of puzzles and more by training dynamics: flat weighting and collapsed rollouts can waste compute, and reward alone can be a poor proxy for faithful reasoning. Looking forward, we believe puzzle-based RLVR can be strengthened by expanding to richer families of verifiable puzzles, improving exploration signals for structured outputs, and further developing consistency diagnostics and objectives that anticipate downstream generalization.

%% file: sec/X_supp.tex
\clearpage
\renewcommand{\thesection}{S\arabic{section}}
\renewcommand{\thefigure}{S\arabic{figure}}
\renewcommand{\thetable}{S\arabic{table}}
\renewcommand{\theequation}{S\arabic{equation}}
\renewcommand*{\thesection}{\Alph{section}}
\setcounter{page}{1}
\setcounter{section}{0}
\setcounter{figure}{0}
\setcounter{table}{0}

\section{Additional Discussion and Details on our Puzzles}
\label{supp:discussion_details}

Measuring the intrinsic complexity of a visual puzzle is nontrivial. In practice, we expose a single \emph{difficulty knob} per puzzle type. For \textbf{Rotation}, the knob is the cardinality of the angle set; in our experiments we fix it to $\{0^\circ,90^\circ,180^\circ,270^\circ\}$ and consider standard variations such as clockwise vs.\ counterclockwise phrasing. For \textbf{PatchFit}, the knob is \emph{distractor hardness}: given a ground-truth patch, we sample $D\in\{3,5,7\}$ decoys drawn from mirror/rotation/color perturbations of the true patch or visually similar patches from other regions. For \textbf{Jigsaw}, difficulty is controlled by grid size: given an $M{\times}N$ grid, we allow any integer pair with $2 \le MN \le 9$ (i.e., up to $3{\times}3$). Sampling is uniform over the chosen configurations.

\begin{figure*}[t]
\centering
\includegraphics[width=\linewidth]{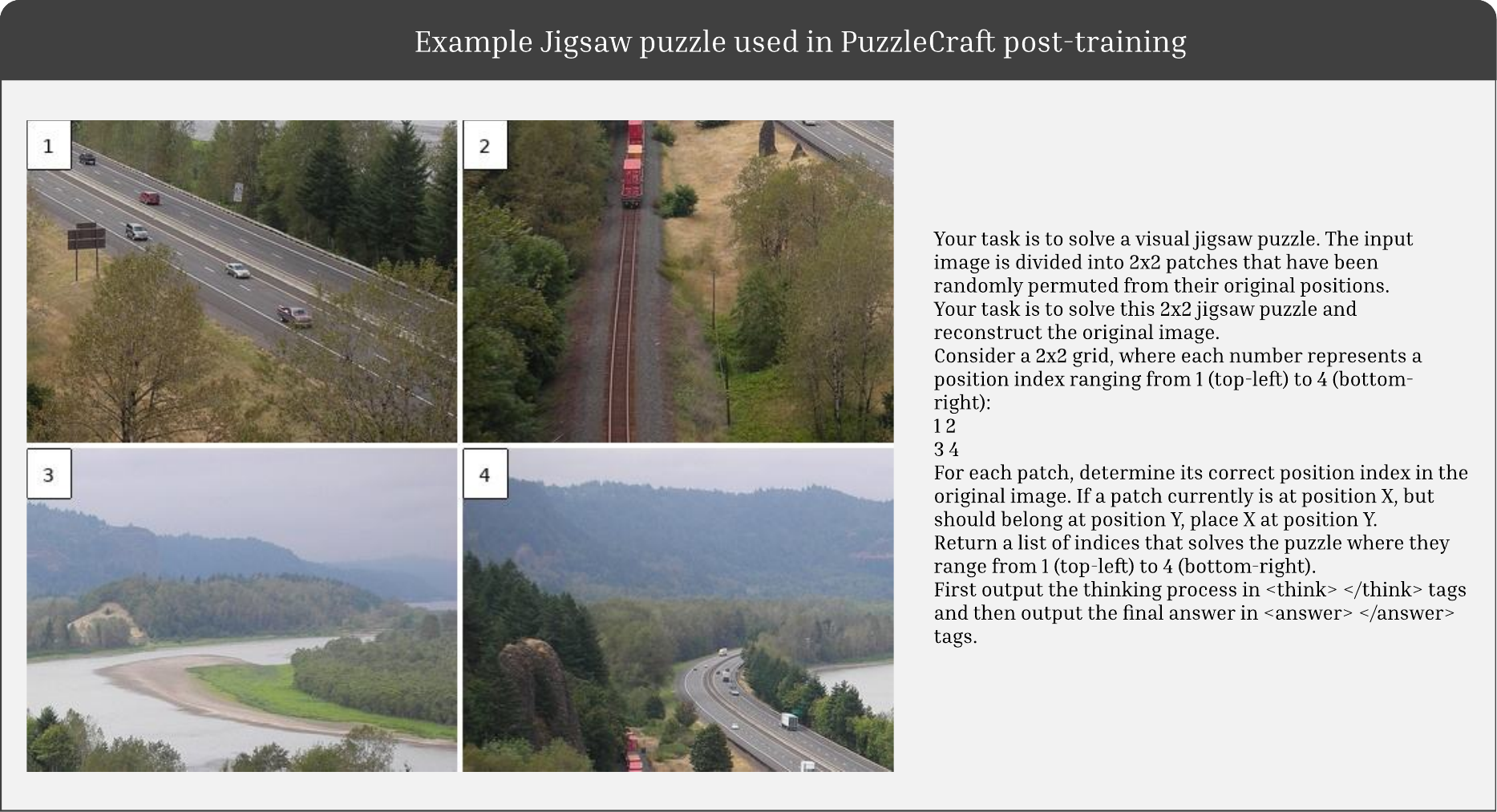}

\caption{Example Jigsaw puzzle used in PuzzleCraft training. The image taken from the Microsoft COCO dataset by Lin et al. is licensed under CC BY 4.0. Source: \url{https://cocodataset.org/}.}

\label{supp:prompts:fig:jigsaw}
\end{figure*}

\begin{figure*}[t]
\centering
\includegraphics[width=\linewidth]{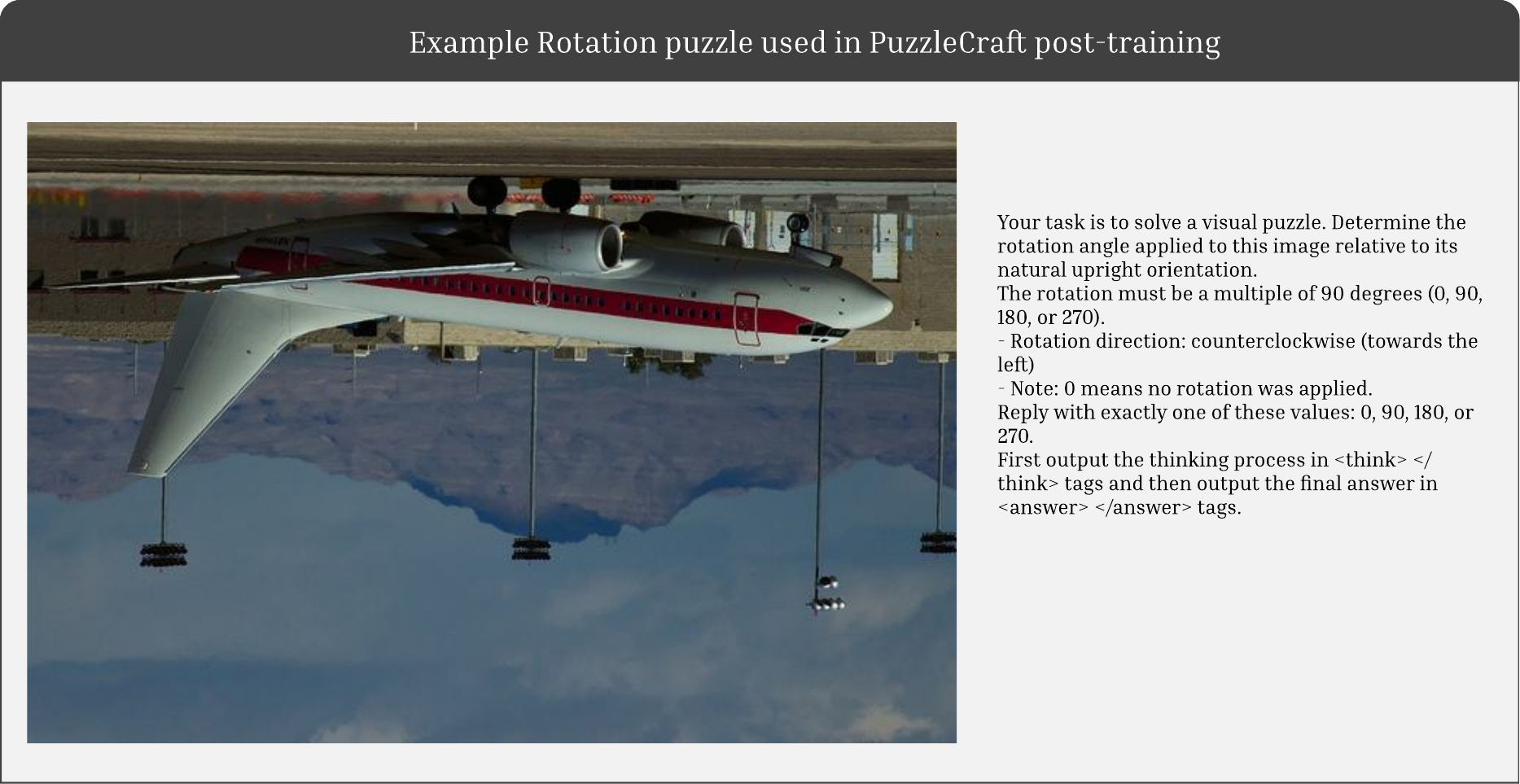}
\caption{Example Rotation puzzle used in PuzzleCraft training. The image taken from the Microsoft COCO dataset by Lin et al. is licensed under CC BY 4.0. Source: \url{https://cocodataset.org/}.}
\label{supp:prompts:fig:rotation}
\end{figure*}

\begin{figure*}[t]
\centering
\includegraphics[width=\linewidth]{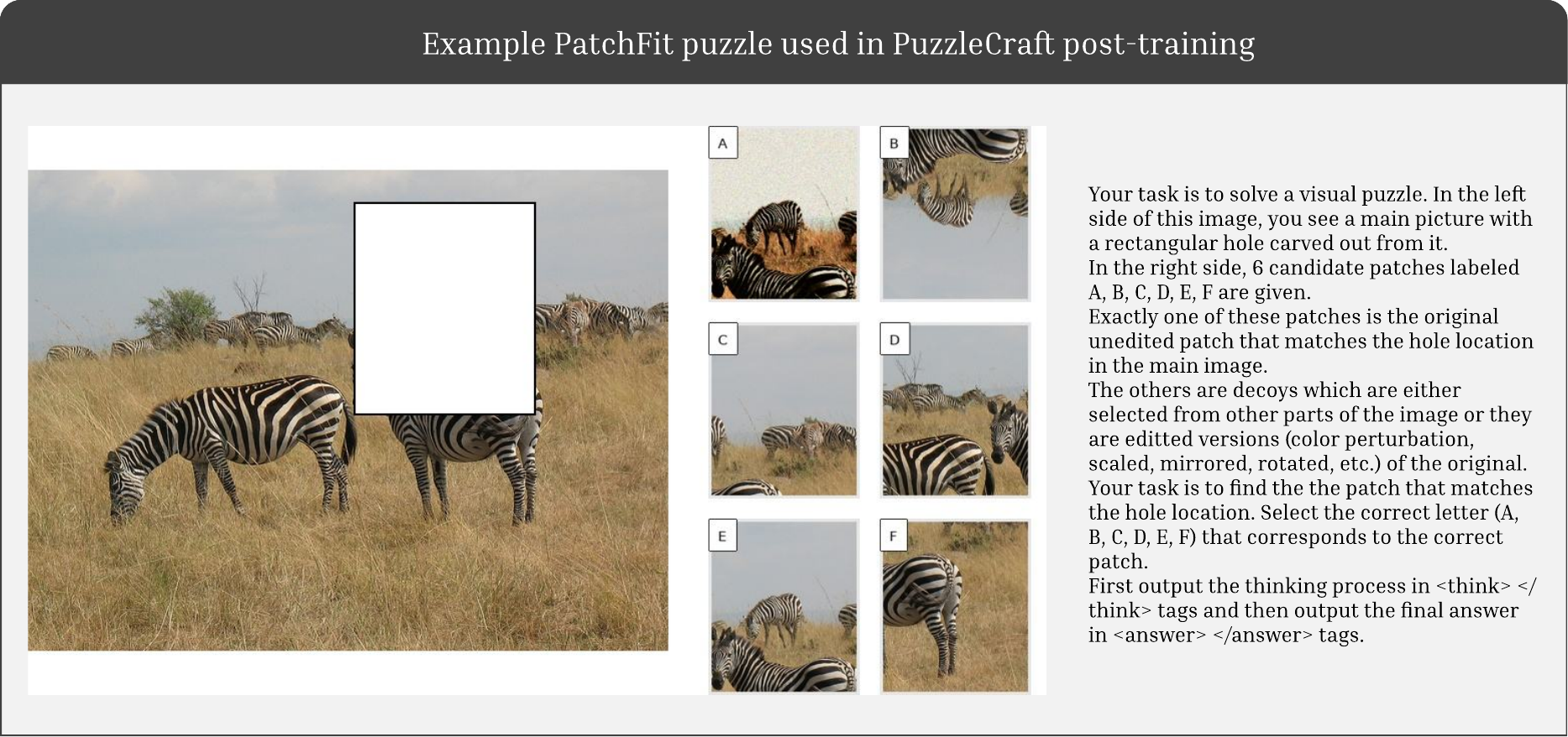}

\caption{Example PatchFit puzzle used in PuzzleCraft training. The image taken from the Microsoft COCO dataset by Lin et al. is licensed under CC BY 4.0. Source: \url{https://cocodataset.org/}.}

\label{supp:prompts:fig:patchfit}
\end{figure*}

Under random guessing, the success rates are as follows. \textbf{Rotation}: $1/4 = 25\%$. \textbf{PatchFit}: averaging over $D\in\{3,5,7\}$ decoys yields an expected success of $\tfrac{1}{4},\tfrac{1}{6},\tfrac{1}{8}$ respectively, i.e., $\approx 18\%$ on average. \textbf{Jigsaw}: with graded reward defined as the fraction of tiles placed correctly, the expected score under a random permutation depends on $MN$; in our sampling it averages to $\approx 26\%$ (grid-size dependent). Exact puzzle-generation details will be provided in the released code. \autoref{supp:prompts:fig:jigsaw} shows a Jigsaw training sample, \autoref{supp:prompts:fig:rotation} a Rotation sample, and \autoref{supp:prompts:fig:patchfit} a PatchFit sample.

\section{Additional Ablations}
\label{supp:sec:curricula}

\input{assets/tables/supp_ablations}

\subsection{Difficulty-Only Curricula}
We compare PuzzleCraft's exploration-aware curriculum against the difficulty-only curriculum of Observe-R1~\cite{guo2025observe-r1} and VCRL~\cite{jiang2025vcrlvariancebasedcurriculumreinforcement}. Results are shown in \autoref{supp:tab:ablations}. Our curriculum consistently performs better, improving Avg. by more than 2.5 points over both methods. This suggests that reward statistics alone are not sufficient for puzzle-based RLVR: accounting for rollout exploration and collapse leads to stronger learning signal and better downstream performance.

\subsection{Frozen Vision Encoder}
We also study a setting where the vision encoder is frozen during post-training, with results reported in \autoref{supp:tab:ablations}. Even in this constrained setup, PuzzleCraft remains strong and clearly outperforms the base model, indicating that a substantial portion of the gains comes from improving the reasoning policy in the language model. At the same time, the full model still performs better than the frozen-ViT variant, suggesting that allowing the visual encoder to adapt provides additional benefits and that part of the improvement also comes from refining the perception module.

\section{RAC Measurement}
\label{supp::sec:rac_exp_setup}

\begin{figure*}[t]
\centering
\includegraphics[width=\linewidth]{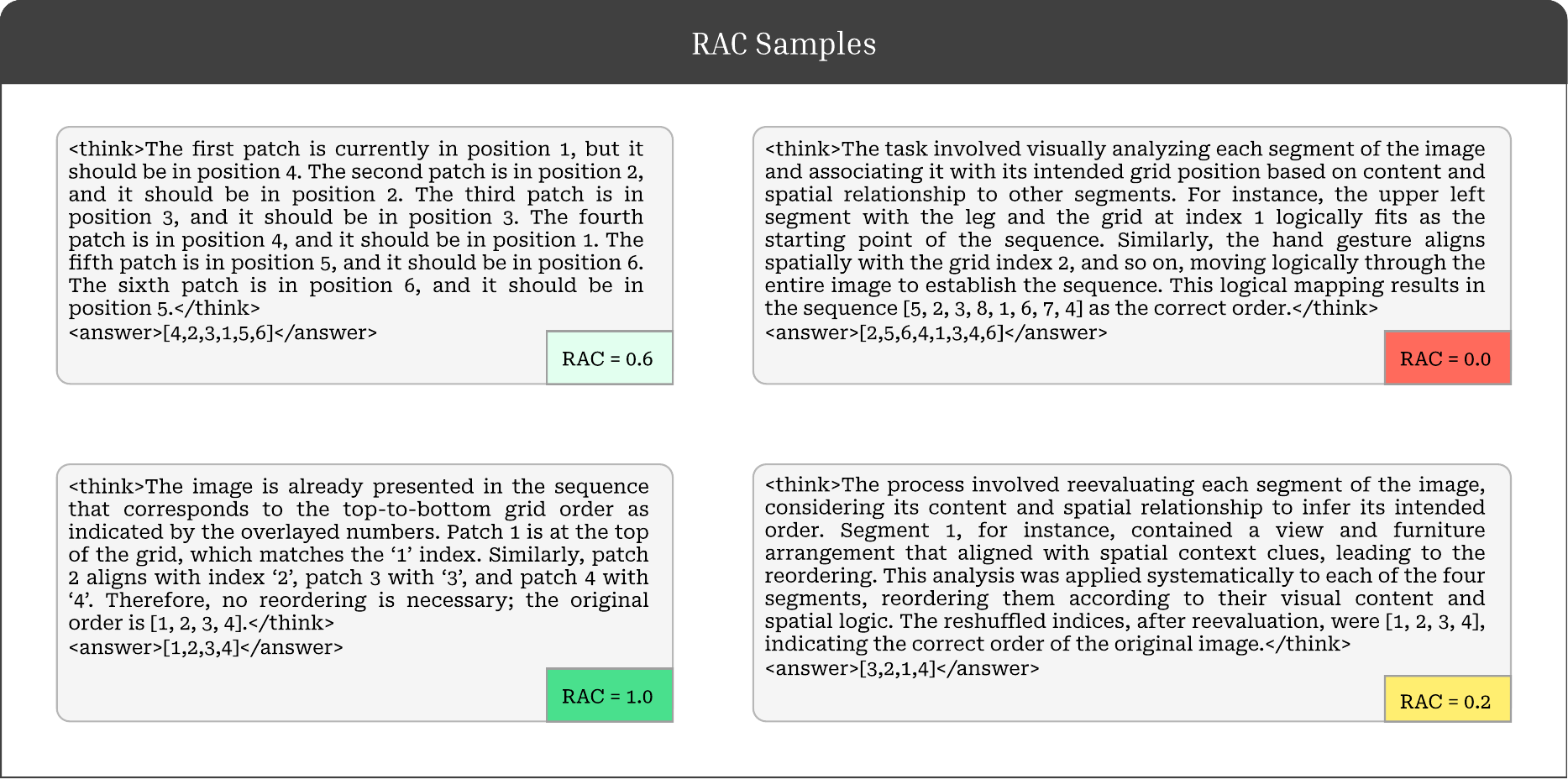}
\vspace{-1em}
\caption{Examples of RAC evaluation during Jigsaw post-training. Each sample shows a model-generated reasoning trace and final answer, together with the RAC score assigned by the judge. These examples illustrate both faithful and inconsistent cases captured by the metric.}
\label{fig:rac_samples}
\vspace{-1em}
\end{figure*}

\begin{figure*}[p!]
\centering
\includegraphics[width=\linewidth]{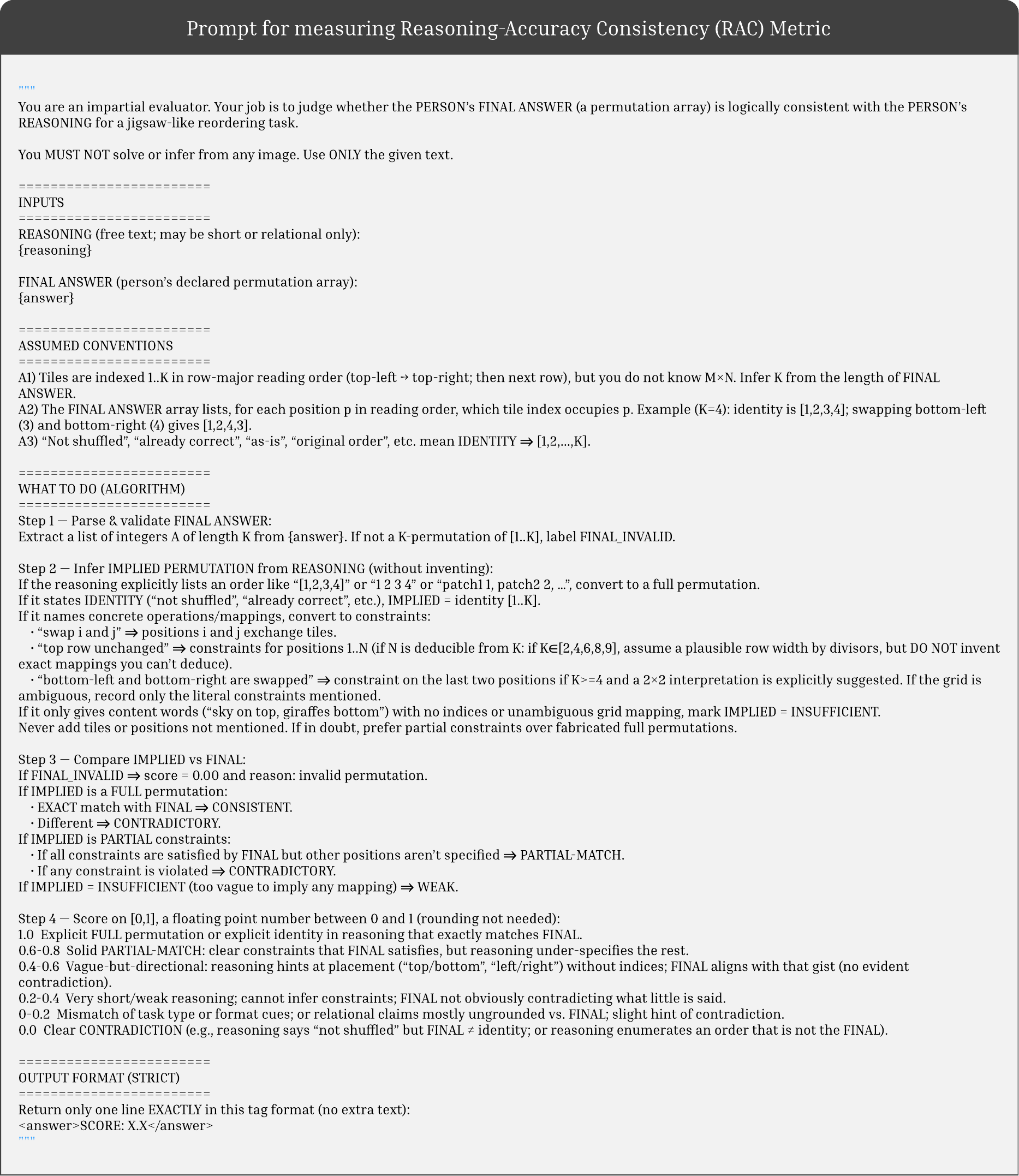}
\caption{Prompt used to measure Reasoning--Answer Consistency (RAC).}
\label{supp:prompts:fig:rac_metric}
\end{figure*}

To measure RAC, we sample rollouts uniformly across the post-training trajectory and, at regular intervals, query a fixed open-source judge (Qwen2.5-VL-72B in inference mode) with each rollout's rationale and final \texttt{<answer>}. The judge determines whether the rationale explicitly supports the answer, and each sample is assigned a binary score in $[0,1]$. The prompt used for this evaluation is shown in \autoref{supp:prompts:fig:rac_metric}. We also provide representative examples from Jigsaw post-training, together with their corresponding RAC scores, to illustrate the kinds of reasoning--answer patterns captured by this metric (\autoref{fig:rac_samples}).

\vspace{0.4em}
\noindent\textbf{Observed pattern.}
Figure~\ref{fig:reasoning_graph} summarizes RAC dynamics for vanilla GRPO, GRPO+curriculum, GRPO+CARE, and GRPO+curriculum+CARE on Jigsaw. Consistent with observations in LLM post-training~\cite{chen2025reasoningmodelsdontsay}, we observe an early increase in faithfulness followed by a later decline for vanilla GRPO. In our setting, the curriculum mitigates this decline, and CARE further improves RAC throughout training. We use RAC as a diagnostic signal for monitoring consistency dynamics rather than as a strict model-selection criterion.

%% file: assets/tables/supp_ablations.tex
\begin{table*}[t]
\centering
\footnotesize
\renewcommand{\arraystretch}{1.4}

\caption{Additional ablations of PuzzleCraft on image benchmarks. We compare our exploration-aware curriculum against difficulty-only curricula (Observe-R1 and VCRL), and further study a frozen-ViT setting to separate gains from visual adaptation versus improvements in the reasoning policy. Avg. is the mean across all 9 benchmarks, with MME normalized as $(\mathrm{MME}/2800)\times 100$ before averaging.}
\vspace{-0.5em}
\begin{adjustbox}{max width=\textwidth}
    
		\begin{tabular}{l|c c c c c c c c c | c}
			\hline
			\textbf{Design choice} & MathVista & MathVision & MathVerse & MME &	MMStar &	POPE &	MMT &	CV-Bench & MMVP   & Avg.\\
			\hline
			 
			Qwen2.5-VL-Instruct & 66.30 & 23.68 & 56.49 & 2243 & 64.67 & 81.77 & 59.64 & 73.62 & 77.67 & 64.88 \\
            
            Jigsaw w/ CL + CARE (Main) & 68.20 & 30.92 & 56.70  & 2366 & 64.60 & 86.52 & 62.26 & 77.63 & 78.67 & 67.78 \\
            
            \hline

            \rowcolor[HTML]{EFEFEF} \multicolumn{11}{l}{\textbf{Curricula}} \\\hline
            
			Main w/ Observe-R1 & 67.40 & 23.02 & 54.49  & 2263 & 64.86 & 83.59 & 62.04 & 75.54 & 74.33 & 65.12 \\
            
			Main w/ VCRL & 64.00 & 26.31 & 55.99  & 2296 & 62.20 & 83.58 & 60.12 & 74.49 & 78.33 & 65.22 \\

            \hline

            \rowcolor[HTML]{EFEFEF} \multicolumn{11}{l}{\textbf{Frozen ViT}} \\\hline
            
			Main w/  Frozen ViT & 68.00 & 25.32 & 54.00  & 2340 & 65.40 & 85.64 & 61.04 & 76.69 & 78.00 & 66.41 \\

			\hline
		\end{tabular}%
    \end{adjustbox}

 	\label{supp:tab:ablations}

\end{table*}